%% file: main.tex
\documentclass{article} 
\usepackage{iclr2027_conference,times}

\input{math_commands.tex}

\usepackage{hyperref}
\usepackage{url}
\usepackage{booktabs} \usepackage{colortbl}
\usepackage{float}
\usepackage{placeins}
\usepackage{needspace}
\usepackage{graphicx}
\usepackage{enumitem}

\definecolor{cvprblue}{RGB}{0,90,156}
\hypersetup{
  colorlinks=true,
  linkcolor=black,
  citecolor=cvprblue,
  urlcolor=cvprblue,
  pdftitle={PAIR: Bridging Perception and Action in Vision-Language-Action Models},
  pdfauthor={Kaixi Feng, Guoheng Sun, Ang Li}
}

\title{\textbf{\textup{PAIR}}: Bridging Perception and Action\\in Vision--Language--Action Models}

\author{Kaixi Feng \quad Guoheng Sun \quad Ang Li \\
University of Maryland, College Park
}

\iclrfinalcopy
\begin{document}
\raggedbottom

\maketitle
\fancyhead{}
\renewcommand{\headrulewidth}{0pt}

\begin{abstract}
Vision--language--action (VLA) models map visual observations and language instructions to continuous robot actions. This task requires a transition from representations that describe the scene and instruction to representations that support action generation. Many continuous-action VLAs leave this transition implicit and supervise it mainly through the final action-prediction loss. We introduce \textbf{PAIR}, a framework that learns a shared perception--action representation between these two spaces. During training, a Masked Action Autoencoder encodes expert action chunks into horizon-aligned Action Latent Tokens. A Bridge Module extracts task-relevant features from the current visual--language representations. PAIR aligns these features with the Action Latent Tokens to form Bridge Tokens that preserve task information and capture the structure of expert actions. The Bridge Tokens are then projected into the action-token space and injected into the initial Action Tokens, providing an action-ready starting point for Action Expert refinement. At inference, the autoencoder is removed, and the Bridge Tokens are generated only from the current observation and instruction. Experiments on LIBERO, LIBERO-Plus, and CALVIN ABC$\rightarrow$D show gains for the evaluated OpenVLA-OFT and VLA-Adapter models. On LIBERO-Plus, PAIR raises VLA-Adapter's success rate from 59.1\% to 64.2\%. On CALVIN, it increases VLA-Adapter's average completed sequence length from 4.42 to 4.53. Across seven real-world tasks, PAIR raises OpenVLA-OFT's success rate from 51.4\% to 65.0\%. Representation analyses show that Bridge Tokens retain task information while making continuous-action information accessible before Action Expert refinement. These results support a shared intermediate representation as a useful interface between perception and action in continuous-action VLAs.
\end{abstract}


\section{Introduction}
\label{sec:introduction}

Vision-language-action (VLA) policies have emerged as a promising framework for robot control. By mapping visual observations and language instructions to continuous actions, they unify visual grounding, language understanding, and control within a single policy. Recent models have achieved strong performance across diverse manipulation tasks \citep{brohan2022rt1,brohan2023rt2,kim2024openvla,wang2025vlaadapter}. Yet this unified formulation hides a difficult representational transition.

The visual-language perception space and the action space are organized around fundamentally different semantics. The former describes the current world and the intended task, including scene content, object relations, and instruction-level intent. The latter specifies how the robot should act through temporally ordered, executable control. We refer to this semantic mismatch as the \emph{perception--action representation gap}.

Existing VLA policies learn to traverse this gap primarily through end-to-end action supervision. The final action loss specifies which action should be produced, while the representation that connects visual-language semantics to action semantics is left implicit. Action tokens must therefore discover within the policy both what information is relevant to control and how it should be reorganized into an action-ready representation.

Recent work has sought to facilitate this transition by exposing control-relevant structure through visual traces, affordances, visual reasoning, and learned latent actions \citep{zheng2024tracevla,nasiriany2024rtaffordance,zhao2025cotvla,liu2026lara}. These efforts show that making such structure explicit can facilitate action generation. Yet robot demonstrations already pair the two sides of the transition: visual--language context specifies what behavior is required, while expert actions reveal how that behavior is executed. This raises a central question:

\begin{figure}[t]
    \centering
    \includegraphics[width=0.88\linewidth]{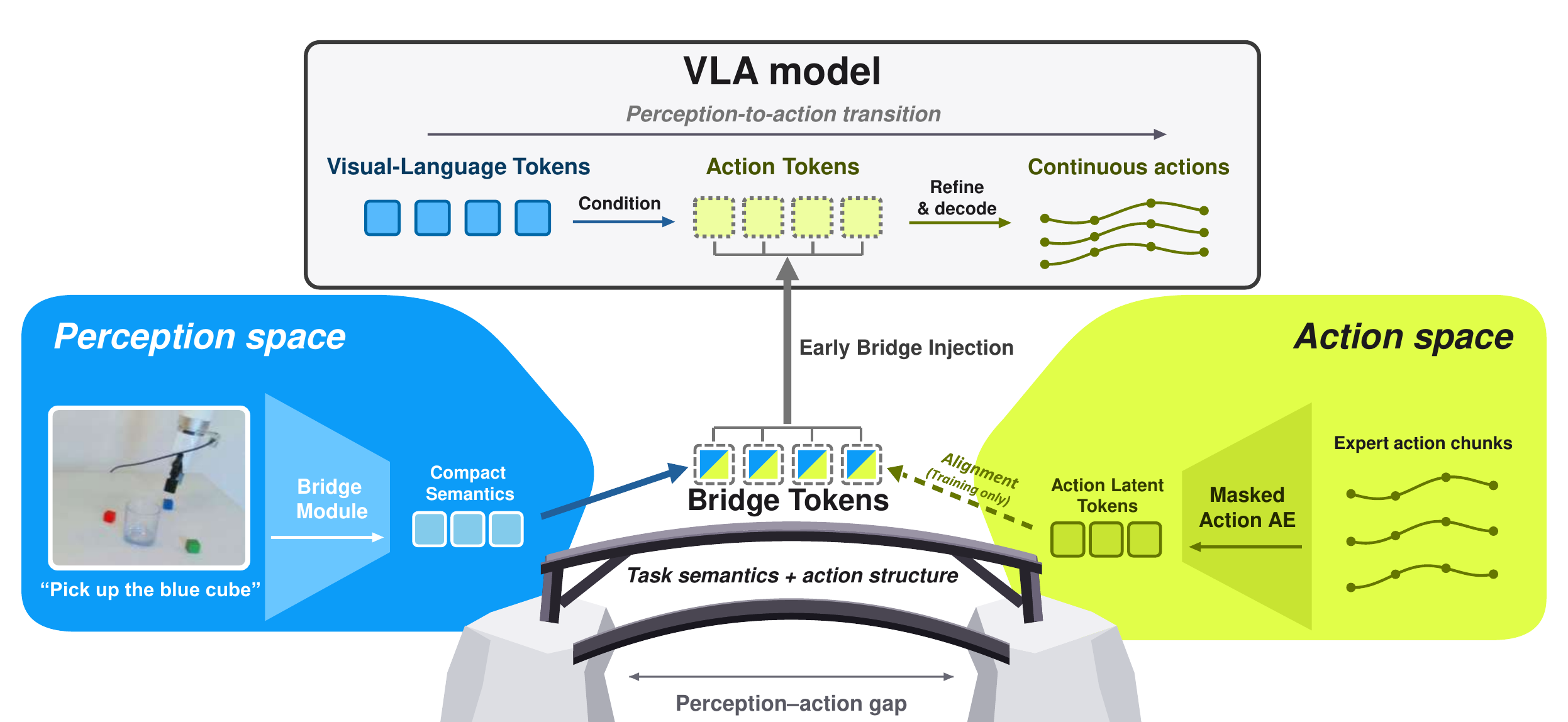}
    \caption{Conceptual view of the perception-to-action transition. Standard VLAs require Action Tokens to acquire and organize action-relevant information during decoding. PAIR instead learns perception-derived Bridge Tokens shaped by the structure of expert actions and injects them into the initial Action Tokens, so that the Action Expert begins from an action-ready intermediate state and refines it toward continuous control. The Action Autoencoder is used only during training.}
    \label{fig:pair-overview}
\end{figure}

\noindent\emph{Can we learn the missing intermediate representation directly from perception and action---a bridge that carries task semantics from one space and executable structure from the other?}

We introduce \textsc{PAIR}, a framework for learning a \emph{Perception--Action Intermediate Representation} directly from paired robot demonstrations. From the action side, a Masked Action Autoencoder encodes expert action chunks into Action Latent Tokens. Reconstructing the demonstrated actions requires these tokens to preserve the temporal and executable structure. From the perception side, a Bridge Module condenses the current visual-language representations into compact, task-relevant features. Aligning these features with the Action Latent Tokens transfers action structure into the perception-derived representation, forming Bridge Tokens that jointly encode task semantics and action information. Perception supplies the semantic content, expert actions supply the executable structure, and alignment integrates both into a unified bridge.

\textsc{PAIR} then projects the Bridge Tokens into the action-token space and injects them at the beginning of action decoding. This early Bridge Injection primes the action tokens with task-aware and action-structured information, providing an explicit starting point for the perception-to-action transition. The policy can then refine this prior through its original action-decoding process rather than constructing an action-ready representation from the final action objective alone.

Experiments on LIBERO \citep{liu2023libero}, LIBERO-Plus \citep{fei2025liberoplus}, and CALVIN ABC$\rightarrow$D \citep{mees2022calvin} demonstrate consistent gains across OpenVLA-OFT and VLA-Adapter. \textsc{PAIR} raises the average success rate of VLA-Adapter from 97.3\% to 98.4\% on LIBERO and from 59.1\% to 64.2\% on LIBERO-Plus. On CALVIN, it improves the average completed sequence length from 4.28 to 4.40 with OpenVLA-OFT and from 4.42 to 4.53 with VLA-Adapter. Across seven real-world tasks, \textsc{PAIR} further raises the average success rate of OpenVLA-OFT from 51.4\% to 65.0\%. These improvements span standard manipulation, zero-shot distribution shifts and long-horizon execution. Representation analyses further show that Bridge Tokens retain task context while exposing action structure before decoding, supporting their role as an explicit interface between perception and action.

\paragraph{Contributions.} Our contributions are as follows:
\enlargethispage{3\baselineskip}
\begin{itemize}[leftmargin=1.5em, labelsep=0.5em, topsep=3pt, itemsep=2pt, parsep=0pt, partopsep=0pt]

\item We characterize the perception--action representation gap in continuous-action VLAs as the implicit transition from visual--language semantics to executable action semantics under final action supervision.

\item We introduce \textsc{PAIR}, which learns perception-derived Bridge Tokens whose task semantics are organized by expert-action structure, and uses early Bridge Injection to provide Action Tokens with an action-aware starting point for refinement.

\item We demonstrate consistent gains across standard manipulation, zero-shot distribution shifts, long-horizon execution, and real-world control, together with representation-level evidence that Bridge Tokens retain task semantics while exposing action structure before Action Expert refinement.

\end{itemize}

\section{Related Work}

\paragraph{Intermediate representations for perception-to-action learning.}
Intermediate representations make control-relevant structure explicit before action prediction. Motion-oriented approaches encode trajectories, visual traces, or future scene changes~\citep{niu2024llarva,zheng2024tracevla,zhang2025dreamvla}. Spatial representations introduce geometric information into perception and action modeling~\citep{qu2025spatialvla,li2025bridgevla}, while affordance-based methods describe possible interactions or key robot configurations~\citep{nasiriany2024rtaffordance,yu2026affordancevla}. These approaches define the intermediate representation through a chosen physical abstraction, such as motion, geometry, or affordance. PAIR does not prescribe a particular physical meaning for the bridge. Its content is learned from the two spaces it is meant to connect: visual--language context provides task semantics, while expert actions organize the representation around executable structure. In this way, task context and action structure jointly determine what the intermediate representation retains.

\paragraph{Latent action representations and tokenization.}
Latent action methods learn compact representations of robot behavior for pretraining, prediction, and control. LAPA derives discrete latent actions from visual transitions for VLA pretraining without robot action labels~\citep{ye2024lapa}, while UniVLA learns task-centric latent actions for policy learning~\citep{bu2025univla}. FAST and VQ-VLA tokenize continuous action chunks for token-based policies~\citep{pertsch2025fast,wang2025vqvla}. These approaches primarily structure how actions themselves are represented, tokenized, or predicted. LARA~\citep{liu2026lara} moves closer to the perception--action interface by aligning latent visual dynamics with intermediate Action Expert features during generation. PAIR targets a different object: a perception-derived intermediate representation that is shaped by expert-action structure during training and made available before Action Expert refinement begins. 

\suppressfloats[t]\begin{figure}[t]
    \centering
    \includegraphics[width=0.75\linewidth]{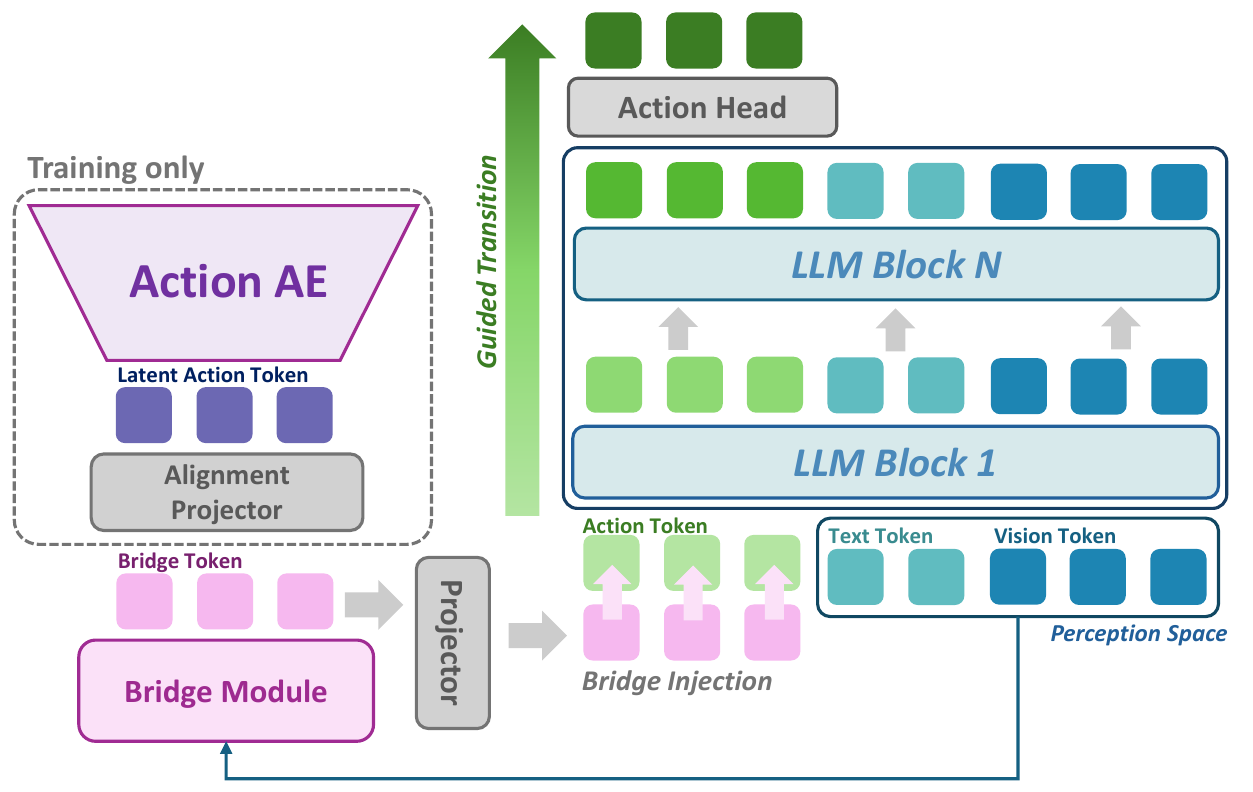}
    \caption{Overview of PAIR. The Masked Action Autoencoder encodes expert actions into Action Latent Tokens, which guide the perception-derived Bridge Tokens through representation alignment. The Bridge Tokens are projected into the Action Token space and injected at the \mbox{entrance of the Action Expert}.}
    \label{fig:pair-framework}
\end{figure}

\section{Method}
\label{sec:method}

\label{sec:method-overview}

We consider a robot demonstration consisting of a visual observation $o$, a language instruction $\ell$, and an $H$-step expert action chunk $\mathbf{A}\in\mathbb{R}^{H\times d_a}$. The VLA backbone encodes the observation and instruction into visual--language representations $\mathbf{H}_{VL}=f_{VL}(o,\ell)$, which condition an Action Expert to predict continuous actions.

A useful representation between perception and action should carry two complementary forms of information: the task semantics that determine what behavior is required, and the executable structure that determines how the behavior should unfold. As illustrated in Figure~\ref{fig:pair-framework}, \textsc{PAIR} learns these properties from the two sides of the transition. A Masked Action Autoencoder captures action structure from expert trajectories, while a Bridge Module extracts compact task-relevant information from the visual--language context. Aligning their horizon-wise representations forms Bridge Tokens, which are injected into Action Tokens at the beginning of the Action Expert.

\enlargethispage{2\baselineskip}
At inference, the Action Autoencoder is removed and the Bridge Module predicts the task-conditioned, horizon-wise tokens directly from the current observation and instruction.

\subsection{Masked Action Autoencoder}
\label{sec:masked-action-ae}

We first construct the action side of the intermediate representation. The desired representation should preserve the information that makes an expert trajectory executable, including its control content and temporal organization. We use trajectory reconstruction as a functional learning objective: a latent sequence that can recover the complete action chunk must retain the structure needed to reproduce that behavior.

The Masked Action Autoencoder receives a corrupted expert trajectory. The corruption operator $\mathcal{C}$ randomly masks action steps and perturbs the sequence with Gaussian noise. This corruption prevents the autoencoder from trivially copying individual action values: masking forces it to recover missing steps from temporal dependencies among action steps, while noise encourages a robust latent representation of the underlying action structure. Its encoder uses self-attention across action steps to model temporal dependencies within the action chunk. The resulting latent sequence is decoded back into the clean trajectory:
\begin{equation}
\widetilde{\mathbf{A}}=\mathcal{C}(\mathbf{A}),
\qquad
\widetilde{\mathbf{Z}}_A
=
E_A(\widetilde{\mathbf{A}}),
\qquad
\widehat{\mathbf{A}}_{\mathrm{AE}}
=
D_A(\widetilde{\mathbf{Z}}_A).
\label{eq:masked-action-ae}
\end{equation}
Here,
$\widetilde{\mathbf{Z}}_A
=
[\widetilde{\mathbf{z}}_{A,1},\ldots,
\widetilde{\mathbf{z}}_{A,H}]
\in\mathbb{R}^{H\times d_z}$
contains one Action Latent Token for each step of \mbox{the action horizon}.

The autoencoder is trained with an $\ell_1$ reconstruction objective:
\begin{equation}
\mathcal{L}_{\mathrm{AE}}
=
\frac{1}{H}
\sum_{i=1}^{H}
\left\lVert
\widehat{\mathbf{a}}_{\mathrm{AE},i}
-
\mathbf{a}_i
\right\rVert_1.
\label{eq:action-ae-loss}
\end{equation}

After this stage, we freeze the autoencoder. During \textsc{PAIR} training, its encoder processes the clean expert action chunk and produces the reference tokens
$\mathbf{Z}_A=\operatorname{sg}[E_A(\mathbf{A})]$.

\subsection{Bridge Module}
\label{sec:bridge-module}

The Action Latent Tokens encode the temporal and control structure of the demonstrated behavior. The visual–language context contains rich scene and task information. The Bridge Module uses learnable queries to extract and organize the subset most relevant to the upcoming action.

The module starts from $H$ learnable queries
$\mathbf{Q}_B=[\mathbf{q}_1,\ldots,\mathbf{q}_H]$.
Through cross-attention, the queries gather task-relevant information from $\mathbf{H}_{VL}$. Self-attention across the resulting slots coordinates this information over the action horizon:
\begin{equation}
\mathbf{B}
=
G_{\phi}
\left(
\mathbf{Q}_B,
\mathbf{H}_{VL}
\right)
\in
\mathbb{R}^{H\times d_b}.
\label{eq:bridge-tokens}
\end{equation}
The learnable queries convert the dense visual–language context into a compact sequence with one slot for each step of the action horizon. At this stage, $\mathbf{B}$ contains task information extracted from the current observation and instruction.

\paragraph{Cross-space alignment.}
The two paths now provide horizon-aligned representations with complementary content: $\mathbf{B}$ captures the task context, while $\mathbf{Z}_A$ captures the structure of the corresponding expert behavior. We connect them through an Alignment Projector and align corresponding \mbox{temporal slots}:
\begin{equation}
\mathcal{L}_{\mathrm{align}}
=
\frac{1}{H}
\sum_{i=1}^{H}
\left[
1-
\cos
\left(
P_{\mathrm{align}}(\mathbf{b}_i),
\mathbf{z}_{A,i}
\right)
\right].
\label{eq:bridge-alignment}
\end{equation}

We use the pre-projector sequence
$\mathbf{B}=[\mathbf{b}_1,\ldots,\mathbf{b}_H]$
as the Bridge Tokens. Their content is selected from the current visual--language context, while the alignment objective organizes that content according to expert action structure. The resulting representation therefore carries task semantics and executable information within the same horizon-wise token sequence.

\subsection{Bridge Injection}
\label{sec:bridge-injection}

Bridge Tokens are learned to mediate the transition from visual--language semantics to executable action. We place them at the beginning of the Action Expert, where Action Tokens enter the computation that develops their action semantics. This early injection primes the Action Tokens with the information represented on both sides of the transition.

Let $\mathbf{X}_A^0\in\mathbb{R}^{H\times d_h}$ denote the Action Tokens entering the Action Expert. A projector maps the Bridge Tokens into the same hidden space, while a token-wise gate regulates their contribution across the action horizon:
\begin{equation}
\widetilde{\mathbf{X}}_A^0
=
\mathbf{X}_A^0
+
\tanh\!\left(G_g(\mathbf{B})\right)
\odot
P(\mathbf{B}).
\label{eq:bridge-injection}
\end{equation}
Here, $P$ projects the Bridge Tokens into the Action Token space, and $G_g$ predicts an input-dependent injection weight for each temporal slot.

The Action Expert processes the primed tokens through its original computation and produces the continuous action chunk. Because task-relevant context and action structure are available from the first expert block, the Action Tokens begin from an explicit intermediate state and refine it \mbox{toward executable control}.

\subsection{Training}
\label{sec:pair-training}

The two training stages follow the construction of the intermediate representation. The first stage learns action structure by optimizing the Masked Action Autoencoder with $\mathcal{L}_{\mathrm{AE}}$. The second stage freezes the autoencoder and transfers this structure to the perception-derived Bridge Tokens while learning to use them inside the policy.

Let $\mathcal{L}_{\mathrm{act}}$ denote the original continuous action prediction loss of the VLA policy. The second-stage objective is
\begin{equation}
\mathcal{L}_{\mathrm{PAIR}}
=
\mathcal{L}_{\mathrm{act}}
+
\lambda_{\mathrm{align}}
\mathcal{L}_{\mathrm{align}},
\label{eq:pair-objective}
\end{equation}
where $\lambda_{\mathrm{align}}$ controls the contribution of the representation alignment objective. The VLA policy and bridge module are optimized jointly in this stage.

During inference, the Bridge Module extracts Bridge Tokens from the current visual--language representations. Bridge Injection then primes the Action Tokens at the beginning of the Action Expert for continuous action prediction.

\begingroup
\setlength{\textfloatsep}{10pt plus 2pt minus 2pt}
\setlength{\floatsep}{8pt plus 2pt minus 2pt}
\setlength{\intextsep}{8pt plus 2pt minus 2pt}
\section{Experiments}\label{sec:experimental-results}
Our experiments address three questions:
\begin{itemize}[leftmargin=1.2em,labelsep=0.4em,topsep=1pt,itemsep=0pt,parsep=0pt,partopsep=0pt]
\item Does PAIR improve manipulation performance, zero-shot generalization, and long-horizon execution, and do these gains extend to real-world tasks? (Secs.~\ref{sec:simulation-results}--\ref{sec:real_world})
\item Do Bridge Tokens encode task semantics and action structure before the Action Expert, and how does their action information compare with baseline Action Tokens across layers? (Section~\ref{sec:representation_analysis})
\item How do action alignment and injection position affect PAIR's performance, and what are the contributions of adaptive injection and autoencoder noise? (Section~\ref{sec:ablations}; Appendix~\ref{app:additional-ablations})
\end{itemize}

\subsection{Experimental Setup}\label{sec:experimental-setup}
\paragraph{Benchmarks and metrics.}
LIBERO~\citep{liu2023libero} evaluates task-conditioned manipulation across the Spatial, Object, Goal, and Long suites, which vary spatial relationships, manipulated objects, task goals, and action-sequence length. We evaluate each task over 50 rollouts from the official initial states and report success rates for each suite and their average. Following the official protocol, LIBERO-Plus~\citep{fei2025liberoplus} evaluates the corresponding LIBERO-trained checkpoints zero-shot under changes in camera viewpoint, robot initial state, language instruction, lighting, background, sensor noise, and object layout. CALVIN ABC$\rightarrow$D~\citep{mees2022calvin} trains policies in environments A--C and evaluates 1,000 five-task chains in unseen environment D. We report success at completing at least one to five consecutive tasks and average completed sequence length, capturing both individual-task performance and sustained execution. Additional protocols are in Appendix~\ref{app:experimental-details}.

\paragraph{Backbones and implementation.}
We evaluate PAIR with VLA-Adapter~\citep{wang2025vlaadapter} on LIBERO/Plus, both VLA-Adapter and OpenVLA-OFT~\citep{kim2025openvlaoft} on CALVIN, and OpenVLA-OFT on the real robot. Each \textbf{+ PAIR} row augments the preceding backbone. We train on four NVIDIA RTX 6000 Ada Generation GPUs; detailed training settings are provided \mbox{in Appendix~\ref{app:experimental-details}}.

\FloatBarrier
\subsection{Simulation Results}\label{sec:simulation-results}
We report PAIR's aggregate results as mean $\pm$ standard deviation over three random seeds.
\vspace{-0.9em}
\paragraph{LIBERO: task-conditioned manipulation.}
\begin{table}[H] \caption{Simulation results on LIBERO. Success rate (\%) is reported for each task suite and their average. Best results are in bold.} \label{tab:libero-main} \centering \small \setlength{\tabcolsep}{5pt} \renewcommand{\arraystretch}{0.95} \begin{tabular}{lccccc} \toprule Method & Spatial & Object & Goal & Long & Avg. \\ \midrule SpatialVLA~\citep{qu2025spatialvla} & 88.2 & 89.9 & 78.6 & 55.5 & 78.1 \\ WorldVLA~\citep{cen2025worldvla} & 87.6 & 96.2 & 83.4 & 60.0 & 81.8 \\ $\pi_0$-FAST~\citep{pertsch2025fast} & 96.4 & 96.8 & 88.6 & 60.2 & 85.5 \\ LARA (full)~\citep{liu2026lara} & 88.0 & 92.0 & 88.5 & 86.0 & 88.6 \\ GR00T N1~\citep{bjorck2025gr00t} & 94.4 & 97.6 & 93.0 & 90.6 & 93.9 \\ $\pi_0$~\citep{black2024pi0} & 96.8 & 98.8 & 95.8 & 85.2 & 94.2 \\ UniVLA~\citep{bu2025univla} & 96.5 & 96.8 & 95.6 & 92.0 & 95.2 \\ UnifiedVLA~\citep{wang2025unifiedvla} & 95.4 & 98.8 & 93.6 & 94.0 & 95.5 \\ OpenVLA-OFT~\citep{kim2025openvlaoft} & 97.6 & 98.4 & 97.9 & 94.5 & 97.1 \\ $\pi_{0.5}$-LARA~\citep{liu2026lara} & 99.0 & 98.5 & 99.0 & 94.5 & 97.8 \\ \midrule \textbf{VLA-Adapter~\citep{wang2025vlaadapter}} & 97.8 & 99.2 & 97.2 & 95.0 & 97.3 \\ \rowcolor{black!10}\textbf{+ PAIR} & \textbf{99.2} & \textbf{99.6} & \textbf{98.6} & \textbf{96.2} & \textbf{98.4} \\ \bottomrule \end{tabular} \end{table}
PAIR raises VLA-Adapter's average success from 97.3\% to $(98.4 \pm 0.3)\%$, improving all four suites (Table~\ref{tab:libero-main}). Gains of 1.4 points on Spatial and Goal and 1.2 on Long extend the benefit of the bridge from basic manipulation to longer action sequences. The Object suite also improves from 99.2\% to 99.6\%.

\paragraph{LIBERO-Plus: zero-shot OOD generalization.}
\begin{table}[H] \caption{Zero-shot LIBERO-Plus results with LIBERO-trained checkpoints. Total follows the benchmark's overall score, not the seven-category mean. Best results are in bold.} \label{tab:libero-plus} \centering \small \setlength{\tabcolsep}{2.8pt} \renewcommand{\arraystretch}{0.95} \begin{tabular}{lcccccccc} \toprule Method & Camera & Robot & Language & Light & Background & Noise & Layout & Total \\ \midrule OpenVLA~\citep{kim2024openvla} & 0.8 & 3.5 & 23.0 & 8.1 & 34.8 & 15.2 & 28.5 & 15.6 \\ WorldVLA~\citep{cen2025worldvla} & 0.1 & 27.9 & 41.6 & 43.7 & 17.1 & 10.9 & 38.0 & 25.0 \\ NORA~\citep{hung2025nora} & 2.2 & 37.0 & 65.1 & 45.7 & 58.6 & 12.8 & 62.1 & 39.0 \\ UniVLA~\citep{bu2025univla} & 1.8 & \textbf{46.2} & 69.6 & 69.0 & 81.0 & 21.2 & 31.9 & 42.9 \\ $\pi_0$~\citep{black2024pi0} & 13.8 & 6.0 & 58.8 & \textbf{85.0} & \textbf{81.4} & \textbf{79.0} & 68.9 & 53.6 \\ \midrule \textbf{VLA-Adapter~\citep{wang2025vlaadapter}} & 36.2 & 37.9 & 74.6 & 70.6 & 76.1 & 58.0 & 69.7 & 59.1 \\ \rowcolor{black!10}\textbf{+ PAIR} & \textbf{44.3} & 45.7 & \textbf{75.1} & 75.8 & 80.3 & 64.7 & \textbf{72.6} & \textbf{64.2} \\ \bottomrule \end{tabular} \end{table}
PAIR improves all seven OOD categories, increasing overall success from 59.1\% to $(64.2 \pm 0.4)\%$ (Table~\ref{tab:libero-plus}). The largest gains occur for camera viewpoint (+8.1 points), robot initial state (+7.8), and noise (+6.7), followed by lighting (+5.2), background (+4.2), and layout (+2.9); language perturbations improve by 0.5 points. Comparing the same LIBERO-trained checkpoints across both benchmarks, PAIR gains 1.1 points on LIBERO and 5.1 points under distribution shift. The overall success-rate drop from LIBERO to LIBERO-Plus decreases from 38.2 points for VLA-Adapter to 34.2 points with PAIR.

\paragraph{CALVIN: long-horizon execution.}
PAIR improves sustained multi-task execution with both backbones (Table~\ref{tab:calvin-abcd}). Average sequence length increases from 4.28 to $4.40 \pm 0.04$ for OpenVLA-OFT and from 4.42 to $4.53 \pm 0.03$ for VLA-Adapter; five-task success rises by 3.9 and 3.7 points, respectively. For comparison, the gains in single-task success are 0.7 and 0.1 points. For both backbones, the margin over the baseline grows as more consecutive tasks are required, showing that the benefit is more pronounced in sustained execution than in completing the first task alone.

\begin{table}[H] \caption{Simulation results on CALVIN ABC$\rightarrow$D. We report success rates (\%) for completing at least one to five consecutive tasks and average completed sequence length. Best results are in bold.} \label{tab:calvin-abcd} \centering \small \setlength{\tabcolsep}{4pt} \renewcommand{\arraystretch}{0.95} \begin{tabular}{lcccccc} \toprule Method & 1 & 2 & 3 & 4 & 5 & Avg. Len. \\ \midrule OpenVLA~\citep{kim2024openvla} & 91.3 & 77.8 & 62.0 & 52.1 & 43.5 & 3.27 \\ RoboVLMs~\citep{li2024robovlms} & 98.0 & 93.6 & 85.4 & 77.8 & 70.4 & 4.25 \\ VPP~\citep{hu2024vpp} & 96.5 & 90.9 & 86.6 & 82.0 & 76.9 & 4.33 \\ UnifiedVLA~\citep{wang2025unifiedvla} & 98.9 & 94.8 & 89.0 & 82.8 & 75.1 & 4.41 \\ DreamVLA~\citep{zhang2025dreamvla} & 98.2 & 94.6 & 89.5 & 83.4 & 78.1 & 4.44 \\ NIAF~\citep{liu2026niaf} & 99.7 & 95.9 & 90.6 & 84.8 & 76.4 & 4.47 \\ HARP-VLA~\citep{zhu2026harpvla} & \textbf{99.8} & \textbf{96.7} & 91.3 & 84.4 & 75.9 & 4.48 \\ VLA-Adapter-Pro~\citep{wang2025vlaadapter} & 98.5 & 95.0 & 90.5 & 85.3 & 80.0 & 4.50 \\ LIRA~\citep{zhang2026lira} & 98.8 & 95.3 & 91.0 & 85.8 & \textbf{80.5} & 4.52 \\ \midrule \textbf{OpenVLA-OFT~\citep{kim2025openvlaoft}} & 96.9 & 92.0 & 85.7 & 80.4 & 72.9 & 4.28 \\ \rowcolor{black!10}\textbf{+ PAIR} & 97.6 & 93.1 & 88.8 & 83.7 & 76.8 & 4.40 \\ \midrule \textbf{VLA-Adapter~\citep{wang2025vlaadapter}} & 99.1 & 94.6 & 88.8 & 82.8 & 76.5 & 4.42 \\ \rowcolor{black!10}\textbf{+ PAIR} & 99.2 & 96.1 & \textbf{91.7} & \textbf{86.0} & 80.2 & \textbf{4.53} \\ \bottomrule \end{tabular} \end{table}

\FloatBarrier
\subsection{Real-World Experiments}\label{sec:real_world}
\begin{figure}[!htbp] \centering \includegraphics[width=\linewidth,trim=0 2bp 0 0,clip]{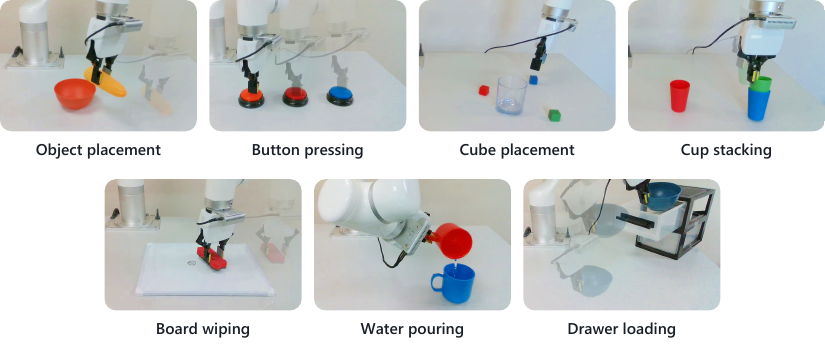} \par\vspace{-5pt} \caption{Seven real-world tasks on the UF850, spanning language grounding, multi-step execution, and precise manipulation.} \label{fig:real_world_tasks} \end{figure}
\paragraph{Platform and tasks.}
We evaluate seven tasks on a UFACTORY UF850 with synchronized third-person and wrist-mounted Intel RealSense RGB observations and robot proprioception (Figure~\ref{fig:real_world_tasks}). Object placement, cube placement, and cup stacking require selecting or arranging objects according to the instruction; button pressing requires executing a specified color order. Board wiping involves sustained tool contact, water pouring requires controlled liquid transfer, and drawer loading combines opening the specified drawer with placing a bowl inside. Together, these tasks cover language-conditioned selection, sequential interaction, precise placement, and contact-rich manipulation. Task instructions and demonstration sequences are provided in Appendix~\ref{app:real-world-details}.

\paragraph{Demonstration data and evaluation.}
We collect expert demonstrations through Quest VR teleoperation, with 30--51 trajectories per task and 293 trajectories in total. OpenVLA-OFT and PAIR use identical processed demonstrations and train a separate policy for each task. Each method is evaluated over 20 trials per task, totaling 140 trials across the seven tasks.

\paragraph{Results.}
PAIR improves OpenVLA-OFT on all seven tasks, from 51.4\% to 65.0\% overall (Figure~\ref{fig:real_world_results}). Gains span object placement (+15 points), sequential button pressing (+5), and drawer loading (+15), with the largest gain in water pouring (+20). Improvements therefore appear in both instruction-driven discrete interactions and tasks requiring extended control over object motion or contact, extending the simulation results to physical manipulation.

\begin{figure}[!htbp]
\centering
\includegraphics[width=0.88\linewidth]{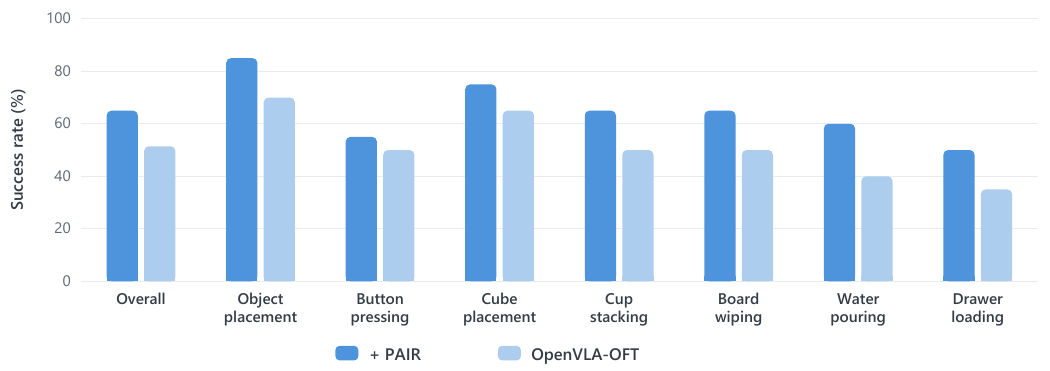}
\par\vspace{-5pt}
\caption{Real-world success rates over 20 trials per task. PAIR improves OpenVLA-OFT across all seven tasks, with an overall gain of 13.6 percentage points.}
\label{fig:real_world_results}
\end{figure}

\FloatBarrier
\input{representation_analysis}

\Needspace{13\baselineskip}
\subsection{Ablation Studies}
\label{sec:ablations}
We ablate the injection position and alignment weight on LIBERO with VLA-Adapter. Each setting is evaluated across all four suites. The action-side target ablation appears in Appendix~\ref{app:additional-ablations}.

\begin{table}[H]
\caption{Injection position ablation on the four LIBERO task suites. Values are success rates (\%).}
\label{tab:injection-position-ablation}
\centering
\small
\setlength{\tabcolsep}{6pt}
\begin{tabular}{lccccc}
\toprule
Position & Spatial & Object & Goal & Long & Avg. \\
\midrule
Start (base) & \textbf{99.2} & \textbf{99.6} & \textbf{98.6} & \textbf{96.2} & \textbf{98.4} \\
End & 98.4 & 99.0 & 97.6 & 94.6 & 97.4 \\
Middle & 98.8 & 98.9 & 98.3 & 95.6 & 97.9 \\
All & 98.7 & 99.0 & 97.9 & 94.8 & 97.6 \\
\bottomrule
\end{tabular}
\end{table}

\paragraph{An early bridge is more effective than repeated injection.}
Entrance injection achieves 98.4\% average success, above end (97.4\%), middle (97.9\%), and all-layer (97.6\%) injection (Table~\ref{tab:injection-position-ablation}). This supports an action-informed starting point for refinement (Section~\ref{sec:representation_analysis}).

\begin{table}[H]
\caption{Alignment loss weight ablation on the four LIBERO task suites. Values are success \mbox{rates (\%)}.}
\label{tab:alignment-weight-ablation}
\centering
\small
\setlength{\tabcolsep}{6pt}
\begin{tabular}{lccccc}
\toprule
$\lambda_{\mathrm{align}}$ & Spatial & Object & Goal & Long & Avg. \\
\midrule
0.1 (base) & \textbf{99.2} & \textbf{99.6} & \textbf{98.6} & \textbf{96.2} & \textbf{98.4} \\
0.01       & 98.8 & 99.2 & 98.0 & 95.2 & 97.8 \\
0          & 98.2 & 98.8 & 97.4 & 94.4 & 97.2 \\
\bottomrule
\end{tabular}
\end{table}

\paragraph{Action alignment strengthens the bridge.}
Without alignment, average success is 97.2\%; $\lambda_{\mathrm{align}}=0.01$ and 0.1 reach 97.8\% and 98.4\% (Table~\ref{tab:alignment-weight-ablation}). Alignment shapes bridge content, while entrance injection sets when it becomes available to the Action Expert.

\FloatBarrier
\endgroup

\section{Conclusion}
\label{sec:conclusion}
PAIR introduces an explicit intermediate representation for the perception-to-action transition in continuous-action VLAs. Bridge Tokens combine visual--language context with action structure and prime Action Tokens at the entrance of the Action Expert. Across LIBERO, LIBERO-Plus, CALVIN, and seven real-world tasks, PAIR improves two VLA backbones in standard manipulation, zero-shot generalization, long-horizon execution, and physical control. Representation analyses show that the learned bridge carries both task and action information, while ablations identify action alignment and early injection as its key ingredients. These results establish explicit perception--action intermediates as a practical design principle for VLA policies.
\section*{AI Use Statement}
Large language models were used to aid language polishing and improve the clarity of the manuscript. All scientific ideas, experimental design, implementation, analysis, results, and conclusions were produced and verified by the authors. We reviewed all AI-assisted text and take responsibility for the final content of this work.

\bibliography{iclr2027_conference}
\bibliographystyle{iclr2027_conference}

\appendix
\setlength{\textfloatsep}{10pt plus 2pt minus 2pt}
\setlength{\floatsep}{8pt plus 2pt minus 2pt}
\setlength{\intextsep}{8pt plus 2pt minus 2pt}
\section{Additional Ablations}
\label{app:additional-ablations}
We report further comparisons on the four LIBERO suites using VLA-Adapter. The base setting is the same as in the main-text ablations; all table entries are success rates (\%). \begin{table}[H] \caption{Action-side alignment target ablation on LIBERO. Success rates (\%).} \label{tab:action-target-ablation} \centering \small \setlength{\tabcolsep}{6pt} \begin{tabular}{lccccc} \toprule Alignment Target & Spatial & Object & Goal & Long & Avg. \\ \midrule Raw expert actions & 98.2 & 98.6 & 97.0 & 94.4 & 97.1 \\ Action Latent Tokens (base) & \textbf{99.2} & \textbf{99.6} & \textbf{98.6} & \textbf{96.2} & \textbf{98.4} \\ \bottomrule \end{tabular} \end{table} \paragraph{Learned action tokens are a stronger alignment target.} The raw-action variant aligns the bridge directly to expert action chunks, whereas PAIR uses the Action Autoencoder's horizon-aligned Action Latent Tokens. The latter reaches 98.4\% average success versus 97.1\% for raw actions (Table~\ref{tab:action-target-ablation}), improving all four suites; its largest gain is 1.8 points on Long.
\begingroup
\setlength{\textfloatsep}{10pt plus 2pt minus 2pt}
\setlength{\intextsep}{8pt plus 2pt minus 2pt}

\begin{table}[H]
\caption{Injection strength ablation on the four LIBERO task suites. Values are success rates (\%).}
\label{tab:injection-strength-ablation}
\centering
\small
\setlength{\tabcolsep}{6pt}
\begin{tabular}{lccccc}
\toprule
Injection Strength & Spatial & Object & Goal & Long & Avg. \\
\midrule
Learnable (base) & \textbf{99.2} & \textbf{99.6} & \textbf{98.6} & \textbf{96.2} & \textbf{98.4} \\
1.0              & 98.8 & 99.4 & 98.2 & 95.6 & 98.0 \\
0.5              & 98.9 & 99.3 & 98.1 & 95.9 & 98.1 \\
\bottomrule
\end{tabular}
\end{table}

\paragraph{Adaptive weighting improves over a fixed contribution.}
Learnable injection reaches 98.4\%, versus 98.0\% and 98.1\% for fixed strengths of 1.0 and 0.5 (Table~\ref{tab:injection-strength-ablation}), supporting adaptive \mbox{bridge weighting}.

\begin{table}[H]
\caption{Action AE noise ablation on the four LIBERO task suites. Values are success rates (\%).}
\label{tab:ae-noise-ablation}
\centering
\small
\setlength{\tabcolsep}{6pt}
\begin{tabular}{lccccc}
\toprule
AE Training & Spatial & Object & Goal & Long & Avg. \\
\midrule
Noise (base) & \textbf{99.2} & \textbf{99.6} & \textbf{98.6} & \textbf{96.2} & \textbf{98.4} \\
Clean        & 98.4 & 99.0 & 97.6 & 95.4 & 97.6 \\
\bottomrule
\end{tabular}
\end{table}

\paragraph{Noise during autoencoder training.}
Training with noise improves average success from 97.6\% to 98.4\% and improves every suite. Denoising therefore produces action structure that remains useful under input corruption.

\FloatBarrier
\endgroup
\input{appendix_analysis}

\Needspace{6\baselineskip}
\section{Implementation and Evaluation Details}
\label{app:experimental-details}

\subsection{Training Hyperparameters}
\begin{table}[H]
\caption{Default training hyperparameters. Batch denotes the effective batch size across four GPUs. The OpenVLA-OFT learning rate remains constant because its decay milestone follows \mbox{the 50k-step run}.}
\label{tab:training-hyperparameters}
\centering
\small
\setlength{\tabcolsep}{4.5pt}
\renewcommand{\arraystretch}{0.96}
\begin{tabular}{lccccc}
\toprule
Stage / backbone & Batch & Steps & Learning rate & Schedule & LoRA rank \\
\midrule
Action Autoencoder & 64 & 50k & $3\!\times\!10^{-4}$ & cosine, 1k warmup & -- \\
VLA-Adapter + PAIR & 64 & 150k & $2\!\times\!10^{-4}$ & cosine & 64 \\
OpenVLA-OFT + PAIR & 64 & 50k & $5\!\times\!10^{-4}$ & constant & 32 \\
\bottomrule
\end{tabular}
\end{table}

\paragraph{Additional settings.}
The Action Autoencoder is trained for 50k steps on LIBERO demonstrations for LIBERO and zero-shot LIBERO-Plus, on CALVIN demonstrations for CALVIN, and separately on each task's demonstrations for the seven real-world tasks. We use AdamW for both stages. The Action Autoencoder uses weight decay $10^{-4}$ and clips the gradient norm at 1.0. Policy training uses bfloat16, image augmentation, and alignment weight $\lambda_{\mathrm{align}}=0.1$.

\paragraph{PAIR architecture.}
The Action Autoencoder comprises a two-layer pre-norm Transformer encoder and a one-layer pre-norm Transformer decoder, both using self-attention over the action chunk with four attention heads. During training, action steps are randomly masked with probability 0.5 and Gaussian noise with standard deviation 0.05 is added before reconstruction. Both OpenVLA-OFT and VLA-Adapter use eight Bridge Tokens, matching the eight action-step tokens at the input to their final Action Experts. The Bridge Module forms them from learned horizon-aligned queries using one pre-norm cross-attention block over the visual--language tokens, followed by one pre-norm self-attention block; both use eight attention heads and residual MLP blocks. The alignment and injection projections are two-layer MLPs with GELU activations. Injection is scaled by a zero-initialized per-token gate implemented by layer normalization, a scalar linear projection, and \mbox{a $\tanh$ nonlinearity}.

\subsection{Simulation Benchmarks}
\paragraph{LIBERO.}
The Action Autoencoder is trained on LIBERO data, and the VLA-Adapter policy is trained on the corresponding suite. LIBERO~\citep{liu2023libero} evaluates task-conditioned manipulation across the Spatial, Object, Goal, and Long suites. These suites vary spatial relationships, manipulated objects, task goals, and the length of the required action sequence. We evaluate each task for 50 rollouts from the official initial states using 256$\times$256 environment observations, and report each suite's success rate and their average.

\paragraph{LIBERO-Plus.}
Following the official evaluation protocol~\citep{fei2025liberoplus}, we directly evaluate the corresponding LIBERO-trained checkpoints zero-shot. The seven perturbation categories cover camera viewpoint, robot initial state, language instruction, lighting, background, sensor noise, and object layout. Per-category and overall success assess generalization beyond the \mbox{original training conditions}.

\paragraph{CALVIN.}
We use the ABC$\rightarrow$D split~\citep{mees2022calvin}: policies train in environments A--C and execute 1,000 five-task chains in unseen environment D. OpenVLA-OFT executes each eight-step action chunk open-loop before replanning. We report success at completing at least one to five consecutive tasks and average completed sequence length, measuring both individual-task performance and sustained execution.

\subsection{Real-World Tasks and Evaluation}
\label{app:real-world-details}

\paragraph{Platform, data, and evaluation.}
All physical experiments use a UFACTORY UF850 arm with synchronized wrist-mounted and fixed third-person Intel RealSense RGB cameras. The policy receives both RGB views and robot proprioception. Expert demonstrations are collected through Quest VR teleoperation and converted to RLDS. For each task, we train a separate Action Autoencoder for PAIR and separate policies for OpenVLA-OFT and PAIR on identical processed trajectories. We evaluate 20 trials per task (140 in total) and report the macro-average across the seven tasks.

\begin{table}[!htbp]
\caption{Real-world task definitions and dataset sizes. $X$, $Y$, and $Z$ denote instance-specific objects or colors.}
\label{tab:real-world-task-definitions}
\centering
\small
\setlength{\tabcolsep}{4pt}
\renewcommand{\arraystretch}{1.04}
\begin{tabular}{@{}p{0.18\linewidth}p{0.67\linewidth}c@{}}
\toprule
Task & Instruction and success criterion & Demos. \\
\midrule
Object placement & ``Place the object in the $Y$ bowl.'' Place the object in the instructed bowl. & 51 \\
Button pressing & ``First press $X$, then $Y$, and finally $Z$.'' Press the three buttons in the specified order. & 50 \\
Cube placement & ``Put $X$ in the cup.'' Select the instructed cube among the candidates and place it in the cup. & 51 \\
Cup stacking & ``Stack $X$ on top of $Y$.'' Stack the instructed source cup on the target cup. & 51 \\
Board wiping & ``Wipe the marked circle off the whiteboard.'' Grasp the eraser and remove the marked circle. & 30 \\
Water pouring & ``Pour water from the red cup into the blue cup.'' Transfer the water between the specified cups. & 30 \\
Drawer loading & ``Open the second drawer and place the blue bowl inside.'' Open the specified drawer and place the bowl inside. & 30 \\
\bottomrule
\end{tabular}
\end{table}

\paragraph{Task coverage.}
The task suite spans instruction-conditioned target selection in object and cube placement, relative object reasoning and terminal alignment in cup stacking, ordered execution in button pressing, and contact-rich or articulated manipulation in board wiping, water pouring, and drawer loading. Together, these tasks cover discrete task specification, sequential control, and extended physical interaction under a shared hardware and observation setup.

\begin{figure}[H]
\vspace*{-6pt}
\centering
\includegraphics[width=0.98\linewidth]{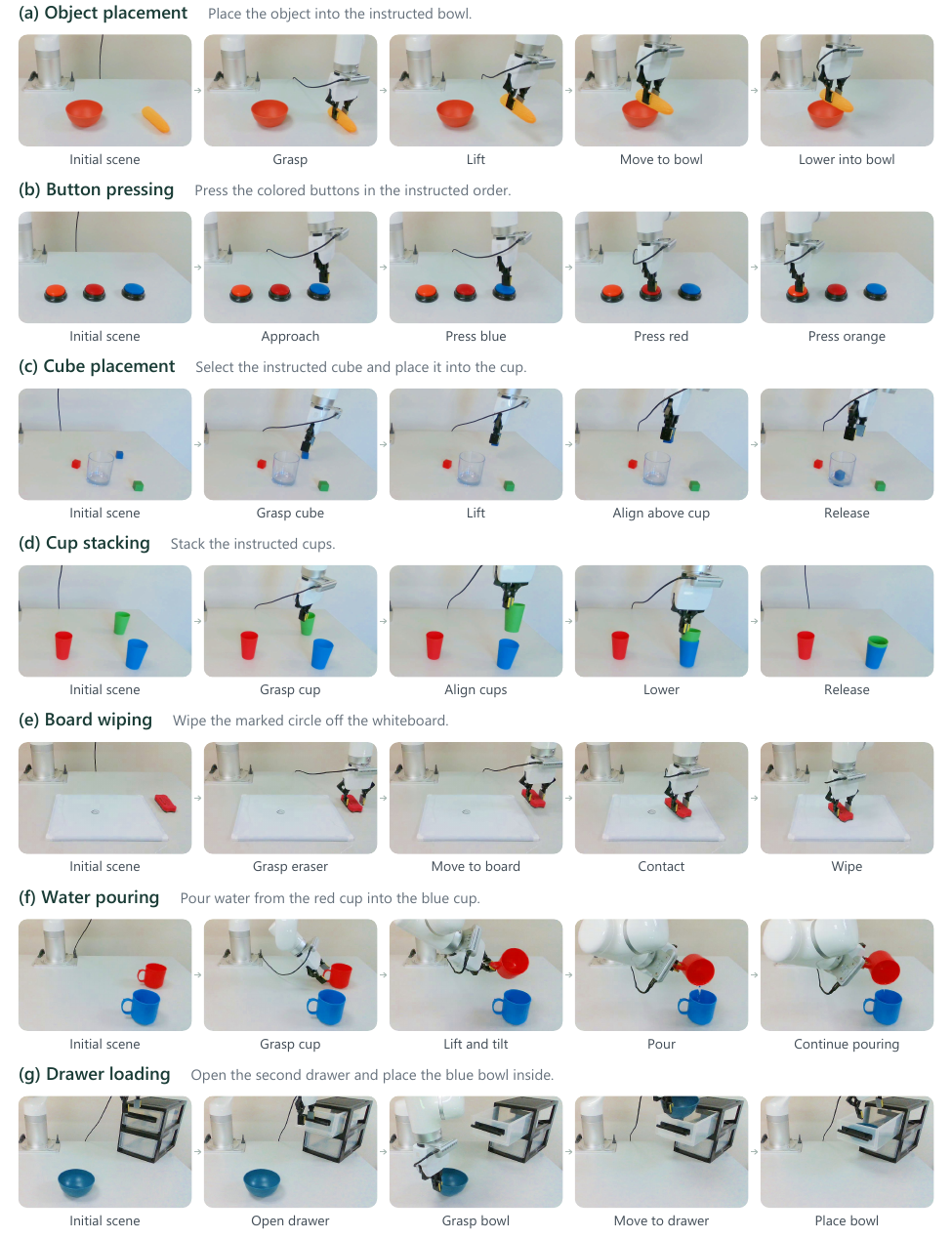}
\caption{\textbf{Demonstration sequences for the seven real-world tasks.} Each row shows five chronological frames from one recording, with short labels identifying the manipulation stages.}
\label{fig:real-world-sequences}
\end{figure}

\end{document}

%% file: math_commands.tex
\usepackage{amsmath,amsfonts,bm}

\def\eqref#1{equation~\ref{#1}}

\def\1{\bm{1}}

\DeclareMathAlphabet{\mathsfit}{\encodingdefault}{\sfdefault}{m}{sl}
\SetMathAlphabet{\mathsfit}{bold}{\encodingdefault}{\sfdefault}{bx}{n}



%% file: representation_analysis.tex
\Needspace{0.40\textheight}
\subsection{Analysis}
\label{sec:representation_analysis}
\begin{figure}[!htbp]
\centering
\includegraphics[width=\linewidth]{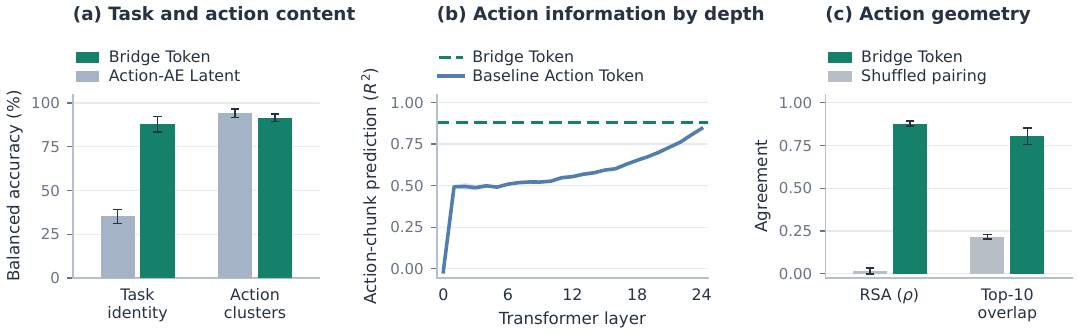}
\vspace{-6pt}
\caption{\textbf{Bridge Tokens connect task semantics to action structure.} (a) Linear-probe balanced accuracy for task identity and action clusters. (b) Action-chunk $R^2$ for baseline Action Tokens across layers (blue) and Bridge Tokens before the Action Expert (dashed). (c) RSA and top-10 neighbor overlap with ground-truth actions. Error bars show fold standard deviations in (a,b), and task standard deviations for Bridge Tokens or 5th--95th permutation percentiles for \mbox{shuffled controls in (c)}.}
\label{fig:bridge_analysis}
\label{fig:bridge_linear_probes}
\label{fig:bridge_geometry}
\vspace{-5pt}
\end{figure}

Using frozen representations from 500 task-balanced LIBERO-Spatial frames, Figure~\ref{fig:bridge_analysis}(a) uses five-fold linear probes with trajectory-grouped splits to predict ten task identities and eight coarse action modes clustered from ground-truth action chunks. It compares Bridge Tokens with latents from the action autoencoder. Figure~\ref{fig:bridge_analysis}(b) decodes the full eight-step, seven-dimensional action chunk from pre-Transformer Bridge Tokens and baseline Action Tokens after each block. Both panels use the same 500 frames and trajectory-grouped folds. Higher accuracy or $R^2$ means more linearly accessible information; details of the layerwise action-decoding and geometry analyses are provided in Appendix~\ref{app:additional-analysis}.

\paragraph{Bridge Tokens encode both task and action information.}
Figure~\ref{fig:bridge_analysis}(a) contrasts task identity with coarse action-mode decodability. Bridge Tokens achieve $87.8\% \pm 4.5\%$ task accuracy versus $35.2\% \pm 4.0\%$ for action-only autoencoder latents. They also retain strong action-cluster accuracy ($91.5\% \pm 2.2\%$), compared with $94.1\% \pm 2.3\%$ for the action-only representation. Bridge Tokens therefore combine substantially richer task information with strongly accessible action structure.

\paragraph{Bridge Tokens make actions readable before Action Expert refinement.}
Baseline Action Tokens progressively acquire continuous-action information through the Transformer, reaching $R^2=0.844$ only at the final block. Bridge Tokens already reach $R^2=0.878$ before the first block (Figure~\ref{fig:bridge_analysis}(b)). Action information is therefore linearly accessible in Bridge Tokens at the entrance of \mbox{the Action Expert}.

\paragraph{The bridge preserves relationships between actions.}
Figure~\ref{fig:bridge_analysis}(c) compares Bridge-token and ground-truth action geometry using cross-trajectory pairs within each task. RSA reaches 0.878 and top-10 neighbor overlap reaches 80.4\%, versus 0.014 and 21.6\% after shuffling. Bridge Tokens therefore preserve continuous action geometry beyond task identity.

%% file: appendix_analysis.tex
\section{Additional Analysis}
\label{app:additional-analysis}
\subsection{Layerwise Probe Protocol}
For the layerwise analysis in Figure~\ref{fig:appendix-depth}, we use frozen representations from LIBERO-Spatial, with 500 training-set frames balanced across the ten tasks. Each example contains eight pre-LLM Bridge Tokens, the corresponding Action Latent Tokens, and an eight-step, seven-dimensional ground-truth action chunk. Both the Bridge--Action Latent and Bridge--baseline Action Token comparisons use the same 500 frames, action targets, and trajectory-grouped folds. Baseline VLA-Adapter Action Tokens are extracted before the Transformer and after each of its 24 blocks.

\paragraph{Classification and regression.}
Representations are flattened, standardized, and reduced to 50 principal components when their input dimension exceeds 128. Preprocessing is fit on the training portion of each fold. Task classification uses the ten task identities; action classification uses eight KMeans clusters of standardized ground-truth action chunks ($K=8$, 20 initializations, seed 7). A multinomial logistic-regression probe with $C=1$ is evaluated using five trajectory-grouped folds, with all frames from each trajectory assigned to the same fold. We report mean balanced accuracy and fold standard deviation. Continuous-action prediction uses ridge regression ($\alpha=10$) on the same trajectory-grouped folds with training-fold target standardization; we report variance-weighted $R^2$. The split seed is 7.

\subsection{Task and Action Information Across Depth}
\begin{figure}[H]
\centering
\includegraphics[width=\linewidth]{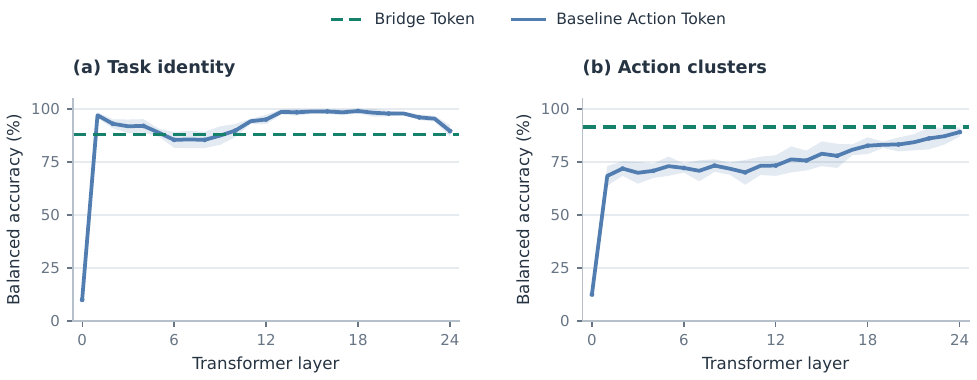}
\caption{\textbf{Task identity and action-cluster decodability across depth.} Blue curves show baseline Action Tokens with fold-standard-deviation bands. Green dashed lines indicate the pre-LLM Bridge Token score. Layer 0 precedes the Transformer; layers 1--24 denote completed blocks.}
\label{fig:appendix-depth}
\end{figure}

Figure~\ref{fig:appendix-depth} separates access to task identity from access to action patterns. Baseline Action Tokens acquire task information rapidly: task accuracy reaches 97.0\% after the first block and remains high through much of the network, ending at 89.6\%. Their action-cluster accuracy develops more gradually and reaches 89.1\% at the final block. Bridge Tokens provide 87.8\% task accuracy together with 91.5\% action-cluster accuracy before entering the Action Expert, making both forms of information accessible from the beginning of action generation.

\paragraph{Depth-wise transition.}
The two baseline curves evolve differently: task identity becomes accessible after the first block, whereas action-cluster structure accumulates more gradually. The Bridge Token operating point combines both signals before any Action Expert block. Its task accuracy remains close to the baseline's late-layer level, while its action-cluster accuracy already exceeds the final baseline value. This pattern complements the entrance-injection ablation: PAIR makes an action-aware state available at the start of refinement, while the baseline develops action structure across later blocks.

\subsection{Continuous-Action Decoding Across Depth}
\begin{figure}[H]
\centering
\includegraphics[width=\linewidth]{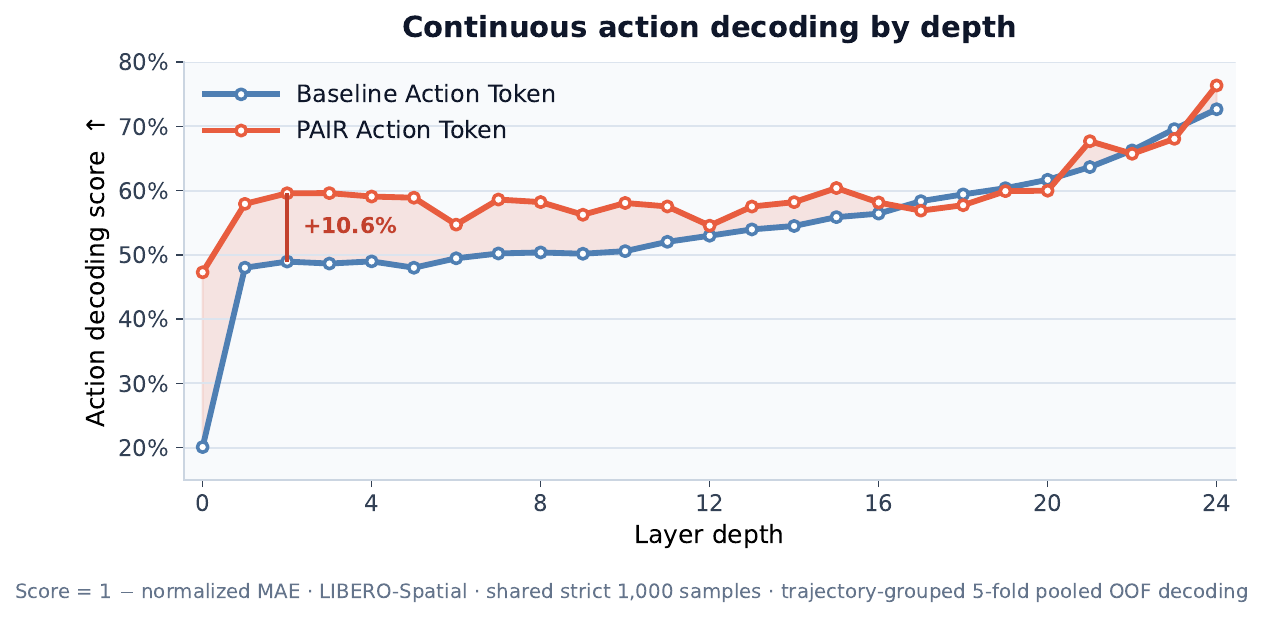}
\caption{\textbf{Continuous-action decodability of Action Tokens across depth.} Ridge readouts predict the full eight-step ground-truth action chunk from frozen Action Tokens. Scores are $1-\mathrm{MAE}$ after target normalization, pooled over five trajectory-grouped held-out folds on 1,000 shared LIBERO-Spatial examples. PAIR tokens are measured after Bridge Token injection; baseline tokens are measured before and after each Transformer block.}
\label{fig:pair-baseline-action-decoding}
\end{figure}

Figure~\ref{fig:pair-baseline-action-decoding} directly compares how much continuous control information is readable from the two Action-Token pathways. After Bridge Token injection, PAIR begins at 47.3\% decoding score, compared with 20.1\% for the baseline tokens before the Transformer. The advantage is largest at the entrance, remains visible across most early and middle depths, and reappears at the final block (76.4\% versus 72.7\%). Thus, the injected Action Tokens start from a more action-readable state and retain that advantage through refinement.

\subsection{Additional Action-Geometry Evidence}
\begin{figure}[H]
\centering
\includegraphics[width=\linewidth]{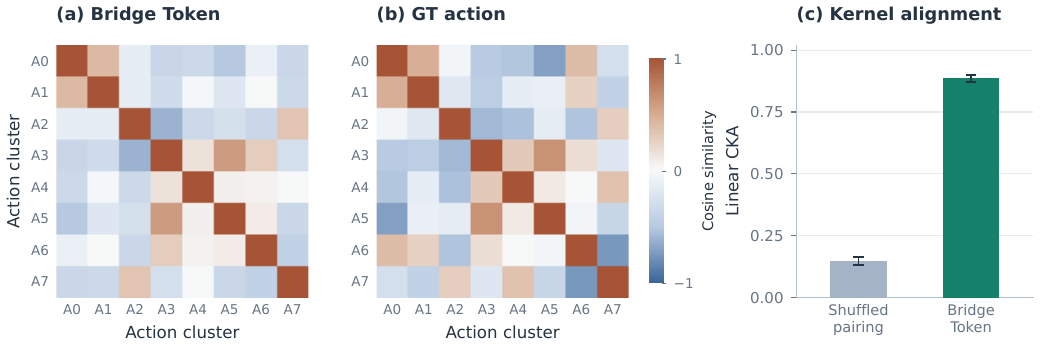}
\caption{\textbf{Action-cluster geometry and kernel alignment.} (a,b) Cosine similarities between GT-derived action-cluster centroids in the two spaces, using a shared color scale. A0--A7 index unsupervised clusters rather than named behaviors. (c) Within-task linear CKA. Error bars indicate task standard deviation for Bridge Tokens and the 5th--95th percentile interval of the trajectory-permutation reference.}
\label{fig:appendix-geometry}
\end{figure}

\Needspace{8\baselineskip}
\paragraph{Sampling and comparison.}
The geometry analysis uses a separate deterministic 500-frame subset: two ordered frames from each of 25 trajectories per task. Bridge representations and ground-truth actions are independently flattened, feature-wise standardized, centered within each task, and L2-normalized. RSA compares cosine-distance ranks for same-task, cross-trajectory pairs. Top-10 neighbor overlap uses the same task and trajectory restrictions. Linear CKA compares the centered cosine Gram matrices within each task. Scores are averaged across the ten tasks. The permutation reference reassigns complete two-frame trajectory groups within each task 200 times while retaining their internal order.

\paragraph{Action geometry across complementary metrics.}
The centroid heatmaps summarize relationships among action-cluster means. Positive similarity between A0 and A1 and between A3 and A5, and negative similarity between A2 and A3, are retained across the two spaces. Differences between the matrices reflect the perceptual and task context carried by Bridge Tokens. At the sample level, linear CKA reaches 0.887 compared with 0.148 under shuffled correspondence, with task-wise values ranging from 0.865 to 0.909. Together with the distance-ranking and neighborhood results in the main text, these metrics show action structure at both sample and cluster levels.
\FloatBarrier

%% file: iclr2027_conference.bib
@inproceedings{kim2024openvla, title={{OpenVLA}: An Open-Source Vision-Language-Action Model}, author={Kim, Moo Jin and Pertsch, Karl and Karamcheti, Siddharth and others}, booktitle={Proceedings of the 8th Conference on Robot Learning}, series={Proceedings of Machine Learning Research}, volume={270}, pages={2679--2713}, year={2025}, publisher={PMLR}}

@inproceedings{zheng2024tracevla, title={{TraceVLA}: Visual Trace Prompting Enhances Spatial-Temporal Awareness for Generalist Robotic Policies}, author={Zheng, Ruijie and Liang, Yongyuan and Huang, Shuaiyi and Gao, Jianfeng and Daum{\'e}, III, Hal and Kolobov, Andrey and Huang, Furong and Yang, Jianwei}, booktitle={International Conference on Learning Representations}, year={2025}}

@article{nasiriany2024rtaffordance,
  title        = {{RT-Affordance}: Affordances are Versatile Intermediate Representations for Robot Manipulation},
  author       = {Nasiriany, Soroush and Kirmani, Sean and Ding, Tianli and Smith, Laura and Zhu, Yuke and Driess, Danny and Sadigh, Dorsa and Xiao, Ted},
  journal      = {arXiv preprint arXiv:2411.02704},
  year         = {2024}
}

@inproceedings{zhao2025cotvla, title={{CoT-VLA}: Visual Chain-of-Thought Reasoning for Vision-Language-Action Models}, author={Zhao, Qingqing and Lu, Yao and Kim, Moo Jin and others}, booktitle={Proceedings of the IEEE/CVF Conference on Computer Vision and Pattern Recognition (CVPR)}, pages={1702--1713}, year={2025}}

@article{liu2026lara,
  title        = {{LARA}: Latent Action Representation Alignment for Vision-Language-Action Models},
  author       = {Liu, Mengya and Jia, Baoxiong and Huang, Jiangyong and Zhang, Jingze and Huang, Siyuan},
  journal      = {arXiv preprint arXiv:2606.07100},
  year         = {2026}
}

@article{brohan2022rt1,
  title        = {{RT-1}: Robotics Transformer for Real-World Control at Scale},
  author       = {Brohan, Anthony and Brown, Noah and Carbajal, Justice and others},
  journal      = {arXiv preprint arXiv:2212.06817},
  year         = {2022}
}

@article{brohan2023rt2,
  title        = {{RT-2}: Vision-Language-Action Models Transfer Web Knowledge to Robotic Control},
  author       = {Brohan, Anthony and Brown, Noah and Carbajal, Justice and others},
  journal      = {arXiv preprint arXiv:2307.15818},
  year         = {2023}
}

@article{wang2025vlaadapter, title={{VLA-Adapter}: An Effective Paradigm for Tiny-Scale Vision-Language-Action Model}, author={Wang, Yihao and Ding, Pengxiang and Li, Lingxiao and others}, journal={Proceedings of the AAAI Conference on Artificial Intelligence}, volume={40}, number={22}, pages={18638--18646}, year={2026}, doi={10.1609/aaai.v40i22.38931}}

@inproceedings{hu2024vpp, title={Video Prediction Policy: A Generalist Robot Policy with Predictive Visual Representations}, author={Hu, Yucheng and Guo, Yanjiang and Wang, Pengchao and others}, booktitle={Proceedings of the 42nd International Conference on Machine Learning}, series={Proceedings of Machine Learning Research}, volume={267}, pages={24328--24346}, year={2025}, publisher={PMLR}}

@article{zhu2026harpvla,  title        = {{HARP-VLA}: Human-Robot Aligned Representation Learning for Vision-Language-Action Model},  author       = {Zhu, Xiang and Yuan, Puzhen and Liu, Yichen and Chen, Jianyu},  journal      = {arXiv preprint arXiv:2605.31234},  year         = {2026}}

@article{zhang2026lira,  title        = {{LIRA}: Local Cross-Layer Information Routing for Vision-Language-Action Decoding},  author       = {Zhang, Zhewei and Wang, Puyue and Qiao, Guanren and others},  journal      = {arXiv preprint arXiv:2608.07596},  year         = {2026}}

@article{bjorck2025gr00t,  title        = {{GR00T N1}: An Open Foundation Model for Generalist Humanoid Robots},  author       = {Bjorck, Johan and Casta{\~n}eda, Fernando and Cherniadev, Nikita and others},  journal      = {arXiv preprint arXiv:2503.14734},  year         = {2025}}

@inproceedings{liu2023libero,
  title     = {{LIBERO}: Benchmarking Knowledge Transfer for Lifelong Robot Learning},
  author    = {Liu, Bo and Zhu, Yifeng and Gao, Chongkai and Feng, Yihao and Liu, Qiang and Zhu, Yuke and Stone, Peter},
  booktitle = {Advances in Neural Information Processing Systems Datasets and Benchmarks Track},
  year      = {2023}
}

@article{mees2022calvin,
  title   = {{CALVIN}: A Benchmark for Language-Conditioned Policy Learning for Long-Horizon Robot Manipulation Tasks},
  author  = {Mees, Oier and Hermann, Lukas and Rosete-Beas, Erick and Burgard, Wolfram},
  journal = {IEEE Robotics and Automation Letters},
  volume  = {7},
  number  = {3},
  pages   = {7327--7334},
  year    = {2022}
}

@inproceedings{niu2024llarva,
  title     = {{LLARVA}: Vision-Action Instruction Tuning Enhances Robot Learning},
  author    = {Niu, Dantong and Sharma, Yuvan and Biamby, Giscard and Quenum, Jerome and Bai, Yutong and Shi, Baifeng and Darrell, Trevor and Herzig, Roei},
  booktitle = {Conference on Robot Learning},
  volume = {270},
  pages = {3333--3355},
  year      = {2025}
}

@inproceedings{zhang2025dreamvla, title={{DreamVLA}: A Vision-Language-Action Model Dreamed with Comprehensive World Knowledge}, author={Zhang, Wenyao and Liu, Hongsi and Qi, Zekun and others}, booktitle={Advances in Neural Information Processing Systems}, volume={38}, year={2025}}

@inproceedings{qu2025spatialvla, title={{SpatialVLA}: Exploring Spatial Representations for Visual-Language-Action Models}, author={Qu, Delin and Song, Haoming and Chen, Qizhi and others}, booktitle={Proceedings of Robotics: Science and Systems}, year={2025}, doi={10.15607/RSS.2025.XXI.011}}

@inproceedings{li2025bridgevla,
  title   = {{BridgeVLA}: Input-Output Alignment for Efficient {3D} Manipulation Learning with Vision-Language Models},
  author  = {Li, Peiyan and Chen, Yixiang and Wu, Hongtao and others},
  booktitle = {Advances in Neural Information Processing Systems},
  volume = {38},
  year    = {2025}
}

@article{yu2026affordancevla,
  title   = {{AffordanceVLA}: A Vision-Language-Action Model Empowering Action Generation through Affordance-Aware Understanding},
  author  = {Yu, Qize and You, Jiadi and Wang, Yuran and others},
  journal = {arXiv preprint arXiv:2606.06155},
  year    = {2026}
}

@inproceedings{ye2024lapa, title={Latent Action Pretraining from Videos}, author={Ye, Seonghyeon and Jang, Joel and Jeon, Byeongguk and others}, booktitle={International Conference on Learning Representations}, year={2025}}

@inproceedings{pertsch2025fast, title={{FAST}: Efficient Action Tokenization for Vision-Language-Action Models}, author={Pertsch, Karl and Stachowicz, Kyle and Ichter, Brian and others}, booktitle={Proceedings of Robotics: Science and Systems}, year={2025}, doi={10.15607/RSS.2025.XXI.012}}

@inproceedings{wang2025vqvla, title={{VQ-VLA}: Improving Vision-Language-Action Models via Scaling Vector-Quantized Action Tokenizers}, author={Wang, Yating and Zhu, Haoyi and Liu, Mingyu and Yang, Jiange and Fang, Hao-Shu and He, Tong}, booktitle={Proceedings of the IEEE/CVF International Conference on Computer Vision (ICCV)}, pages={11089--11099}, year={2025}}

@inproceedings{kim2025openvlaoft, title={Fine-Tuning Vision-Language-Action Models: Optimizing Speed and Success}, author={Kim, Moo Jin and Finn, Chelsea and Liang, Percy}, booktitle={Proceedings of Robotics: Science and Systems}, year={2025}, doi={10.15607/RSS.2025.XXI.017}}

@inproceedings{bu2025univla, title={Learning to Act Anywhere with Task-centric Latent Actions}, author={Bu, Qingwen and Yang, Yanting and Cai, Jisong and Gao, Shenyuan and Ren, Guanghui and Yao, Maoqing and Luo, Ping and Li, Hongyang}, booktitle={Proceedings of Robotics: Science and Systems}, year={2025}, doi={10.15607/RSS.2025.XXI.014}}

@article{cen2025worldvla, title={{WorldVLA}: Towards Autoregressive Action World Model}, author={Cen, Jun and Yu, Chaohui and Yuan, Hangjie and others}, journal={arXiv preprint arXiv:2506.21539}, year={2025}}

@inproceedings{black2024pi0, title={{$\pi_0$}: A Vision-Language-Action Flow Model for General Robot Control}, author={Black, Kevin and Brown, Noah and Driess, Danny and others}, booktitle={Proceedings of Robotics: Science and Systems}, year={2025}, doi={10.15607/RSS.2025.XXI.010}}

@article{hung2025nora, title={{NORA}: A Small Open-Sourced Generalist Vision Language Action Model for Embodied Tasks}, author={Hung, Chia-Yu and Sun, Qi and Hong, Pengfei and Zadeh, Amir and Li, Chuan and Tan, U-Xuan and Majumder, Navonil and Poria, Soujanya}, journal={arXiv preprint arXiv:2504.19854}, year={2025}}

@article{wang2025unifiedvla, title={Unified Vision-Language-Action Model}, author={Wang, Yuqi and Li, Xinghang and Wang, Wenxuan and Zhang, Junbo and Li, Yingyan and Chen, Yuntao and Wang, Xinlong and Zhang, Zhaoxiang}, journal={arXiv preprint arXiv:2506.19850}, year={2025}}

@article{li2024robovlms, title={What Matters in Building Vision-Language-Action Models for Generalist Robots}, author={Li, Xinghang and Li, Peiyan and Qian, Long and others}, journal={Nature Machine Intelligence}, volume={8}, pages={158--172}, year={2026}, doi={10.1038/s42256-025-01168-7}}

@article{liu2026niaf, title={Neural Implicit Action Fields: From Discrete Waypoints to Continuous Functions for Vision-Language-Action Models}, author={Liu, Haoyun and Zhao, Jianzhuang and Chang, Xinyuan and others}, journal={arXiv preprint arXiv:2603.01766}, year={2026}}

@inproceedings{fei2025liberoplus, title={{LIBERO-Plus}: A Progressive Robustness Benchmark for Visual-Language-Action Models}, author={Fei, Senyu and Wang, Siyin and Shi, Junhao and others}, booktitle={Proceedings of the IEEE/CVF Conference on Computer Vision and Pattern Recognition (CVPR)}, pages={38574--38583}, year={2026}}
